\documentclass[conference]{IEEEtran}
\IEEEoverridecommandlockouts
\usepackage{cite}
\usepackage{amsmath,amssymb,amsfonts}
\usepackage{algorithmic}

\usepackage{graphicx}
\graphicspath{{figures/}}
\usepackage{textcomp}
\usepackage{enumitem}
\usepackage{xcolor}
\usepackage{subfigure} 
\def\BibTeX{{\rm B\kern-.05em{\sc i\kern-.025em b}\kern-.08em
    T\kern-.1667em\lower.7ex\hbox{E}\kern-.125emX}}

\usepackage{tabularx}
\usepackage{multirow}
\usepackage{makecell}
\usepackage{tcolorbox}
\tcbuselibrary{breakable}
\tcbuselibrary{listingsutf8}
\usepackage{listings}
\usepackage{courier}
\usepackage{booktabs} 
\usepackage{hyperref} 

\begin{document}

\title{From Simulation to Strategy: \\Automating Personalized Interaction Planning \\for Conversational Agents}

\author{ 
\IEEEauthorblockN{Wen-Yu Chang\textsuperscript{*}\quad Tzu-Hung Huang\textsuperscript{*}\quad Chih-Ho Chen\textsuperscript{*}\quad Yun-Nung Chen}
\IEEEauthorblockA{\textit{Department of Computer Science and Information Engineering}\\
 \textit{National Taiwan University}, Taipei, Taiwan\\
 f10946031@csie.ntu.edu.tw\quad b11902023@ntu.edu.tw\quad b11902155@g.ntu.edu.tw\quad y.v.chen@ieee.org}
 \thanks{\textsuperscript{*} Equal contribution.}

 }

\maketitle

\begin{abstract}
Amid the rapid rise of agentic dialogue models, realistic user-simulator studies are essential for tuning effective conversation strategies. This work investigates a sales-oriented agent that adapts its dialogue based on user profiles spanning age, gender, and occupation. While age and gender influence overall performance, occupation produces the most pronounced differences in conversational intent. Leveraging this insight, we introduce a lightweight, occupation-conditioned strategy that guides the agent to prioritize intents aligned with user preferences, resulting in shorter and more successful dialogues. Our findings highlight the importance of rich simulator profiles and demonstrate how simple persona-informed strategies can enhance the effectiveness of sales-oriented dialogue systems.\footnote{Code \& Prompts: \url{https://github.com/MiuLab/PersonalizedAgent/}}

\end{abstract}

\section{Introduction}

With the ongoing evolution of Agentic AI, researchers have begun to explore its application across diverse domains. Among these, dialogue systems designed for business recommendation tasks have attracted significant attention. Such systems aim to detect latent business intents during conversations and provide timely, relevant suggestions to users.

To enhance system performance, a deeper understanding of user behavior within these interactions is essential. By analyzing patterns in user responses, preferences, and conversational dynamics, it is possible to tailor dialogue strategies that foster engagement and drive successful outcomes. Several studies have investigated how different user profiles influence conversational dynamics and how to design more engaging interactions~\cite{zhang-etal-2018-personalizing, cho2022personalizeddialoguegeneratorimplicit,wang2025knowbettermodelinghumanlike,tseng2024two}. 
For example, Zhang et al.~\cite{zhang-etal-2018-personalizing} were among the first to incorporate user profiles into dialogue systems, enabling the generation of more personalized and engaging user responses. In addition, Cho et al.\cite{cho2022personalizeddialoguegeneratorimplicit} proposed a personalized dialogue generation framework that implicitly detects user persona from dialogue history via conditional variational inference. Their model enables the agent to generate responses that are better aligned with the user’s latent persona, leading to improved engagement and personalization without requiring explicit user profiles. Furthermore, Wang et al.~\cite{wang2025knowbettermodelinghumanlike} introduced USP, a user simulation framework that models implicit user traits—such as personality, speaking style, and goals—directly from dialogue data. By leveraging these latent profiles, their system can generate more human-like and behaviorally consistent user interactions in multi-turn conversations.

In this study, we further extend this line of research by investigating the importance of user behavior modeling in the context of agentic dialogue systems and exploring methodologies for leveraging these insights to improve system efficacy. Specifically, we focus on a sales-oriented dialogue system, \textsc{SalesAgent}, proposed by Chang and Chen~\cite{chang2024injectingsalespersonsdialoguestrategies}, which aims to proactively shift the dialogue topic toward the user’s potential interests and subsequently elicit explicit expressions of interest in the agent’s recommendations. We conduct a thorough analysis of a broad range of user profiles, including gender, age, and occupation, to examine how different user personas impact conversational flow.

Our experimental results show that users with different profiles exhibit significant variations in their interactions with \textsc{SalesAgent}, in terms of success rate and average number of turns. Moreover, we observe that users with different occupations respond differently to the types of topics the agent attempts to pivot toward. Based on these findings, we construct a lightweight sales-oriented dialogue agent framework that integrates the dialogue strategies of \textsc{SalesAgent} with insights from our user behavior analysis, utilizing a vanilla LLM backbone. This framework achieves a higher success rate compared to a baseline agent that lacks awareness of user behavior insights.

Our main contributions are as follows:
\begin{itemize}
\item We present a comprehensive analysis of how user profiles, specifically gender, age, and occupation, impact the conversational dynamics of a sales-oriented agentic dialogue system.
\item We identify significant patterns in how different user personas affect key conversational outcomes such as success rate, and average dialogue length.
\item We propose a lightweight framework that integrates user behavior insights with an LLM-based sales agent, achieving improved performance over baseline models.
\end{itemize}

\section{Related Work}
\subsection{Role-playing Research in LLMs}

The role-playing capabilities of large language models (LLMs) have attracted growing attention in recent research~\cite{park2023generativeagentsinteractivesimulacra, tseng-etal-2024-two, shao2023characterllmtrainableagentroleplaying}.
Park et al.\cite{park2023generativeagentsinteractivesimulacra} introduced Generative Agents with dynamic memory and reflective reasoning to simulate human-like interactions, while Shao et al.\cite{shao2023characterllmtrainableagentroleplaying} proposed Character-LLM, which leverages curated profiles and reconstructed experiences for more consistent persona modeling.
Beyond general frameworks, domain-specific role-playing has been explored in software engineering~\cite{hong2024metagpt, qian2024chatdevcommunicativeagentssoftware, dong2024self}, where MetaGPT~\cite{hong2024metagpt}, ChatDev~\cite{qian2024chatdevcommunicativeagentssoftware}, and self-collaboration frameworks~\cite{dong2024self} enable multi-agent collaboration through structured communication and role-based expertise.
In the medical domain, Wu et al.\cite{wu2023largelanguagemodelsperform} introduced Diagnostic-Reasoning Chain-of-Thought (DR-CoT) prompting for iterative doctor-patient interactions, and Tang et al.\cite{tang2024medagentslargelanguagemodels} proposed MedAgents, a multi-agent framework for collaborative medical reasoning.

\subsection{\textsc{SalesAgent}}

\textsc{SalesAgent}~\cite{chang2024injectingsalespersonsdialoguestrategies} is a dialogue agent designed to bridge the gap between open-domain chitchat and task-oriented dialogues in sales-oriented settings. It builds upon the foundation established by SalesBot~\cite{chiu-etal-2022-salesbot}, which introduced the first large-scale dataset of dialogues that naturally transition from social conversation to task-oriented interactions. SalesBot demonstrated that triggering business opportunities often requires agents to engage users in casual conversation and gradually detect implicit intents. However, early versions of SalesBot suffered from abrupt transitions and limited coherence.

To address these limitations, \textsc{SalesAgent} is trained on an improved dataset, SalesBot 2.0~\cite{chang2023salesbot}, which enhances conversational flow by incorporating longer and smoother transitions and leveraging commonsense knowledge from large language models (LLMs). Inspired by recent advances in prompting paradigms such as chain-of-thought reasoning~\cite{wei2023chainofthoughtpromptingelicitsreasoning} and ReAct prompting paradigms~\cite{yao2023reactsynergizingreasoningacting}, \textsc{SalesAgent} introduces explicit \textit{strategy-guided reasoning} into the dialogue process via chain-of-thought (CoT) prompting. At each turn, it generates two outputs: a \textit{thought}, which articulates its internal reasoning about the conversational state and user intent, and a \textit{response}, which is the surface utterance produced for the user.

The \textit{thought} component adheres to a structured template, capturing four key dialogue strategies:

\begin{itemize}
    \item \textit{The user did not implicitly mention any potential intent; I should continue the chit-chat.}
    \item \textit{The user implicitly mentioned the intent of \texttt{<intent>}; I should smoothly pivot the conversation to the topic of \texttt{<intent>}.}
    \item \textit{The user did not change the topic of \texttt{<intent>}; I should continue the topic.}
    \item \textit{The user has explicitly shown his/her intent of \texttt{<intent>}.}
\end{itemize}

This explicit reasoning allows the agent to maintain contextual coherence, dynamically adjust its strategy, and modulate its level of proactiveness, thereby improving user engagement and reducing conversational aggressiveness.

In addition to CoT-driven strategy control, recent studies have highlighted the importance of modeling user diversity in evaluating sales agents. Cheng et al.~\cite{cheng2025exploringpersonalityawareinteractionssalesperson} proposed \textit{MBTI-based persona simulators} to systematically analyze how different personality traits affect interaction outcomes with \textsc{SalesAgent}. Their experiments showed that user traits such as openness and extraversion influenced both task success rates and conversational dynamics, underscoring the need for adaptable dialogue strategies.

Taken together, \textsc{SalesAgent} represents a step toward \textit{strategically aware, persona-sensitive sales dialogue agents}. Its design emphasizes (1) dynamic intent recognition and topic management, (2) transparent and controllable strategy execution via CoT reasoning, and (3) adaptability to diverse user profiles. These capabilities make \textsc{SalesAgent} a foundational component in our dialogue pipeline, supporting more natural, effective, and user-centered conversational experiences.
\section{Experimental Setup}

Designing and evaluating personalized dialogue systems through human studies can be time-consuming and inefficient, especially when aiming to control for a diverse and balanced distribution of persona attributes such as age, gender, and occupation. Recruiting real users with specific attribute combinations at scale is often impractical. To address this challenge, we leverage large language models (LLMs) to both generate realistic user personas and simulate user behaviors in conversation. This approach enables scalable and controllable experimentation, allowing for systematic analysis of how different persona attributes influence interactions with a sales agent.

\subsection{Persona Attribute Definition}

Each user simulator is assigned a predefined persona composed of four attributes: gender, age, occupation, and personality~\cite{schatzmann2006survey,li2016user}.

Gender is specified as either male or female. Age is categorized into four distinct groups: teen (15--19 years old), adult (20--45 years old), middle-aged (45--65 years old), and elderly (65 years old and above).

Occupations are selected following the International Standard Industrial Classification of All Economic Activities (ISIC). From the full list of 21 sections, we chose six representative sectors and sampled four occupations from each. The selected sectors are: \textit{Agr}(Agriculture), \textit{Info}, \textit{Fin}, \textit{Edu} (Education), \textit{Heal}, and \textit{Arts}. A detailed list of these sectors and sampled occupations is provided in Table~\ref{sec:occ_cate}.

\begin{table}[htbp]
\caption{Selected ISIC Sectors and Sampled Occupations}
\label{sec:occ_cate}
\centering

\resizebox{0.85\linewidth}{!}{%
\begin{tabularx}{\linewidth}{c p{3cm} X}
\toprule
\textbf{Section} & \textbf{Description} & \textbf{Sampled Occupations} \\
\midrule
Agr & Agriculture, Forestry, and Fishing & Farmer, Woodcutter, Fisherman, Horticulturist \\
Info & Information and Communication & Software Engineer, Cybersecurity Specialist, Data Scientist, Telecommunications Technician \\
Fin & Financial and Insurance Activities & Investment Analyst, Actuary, Insurance Claims Adjuster, Financial Advisor \\
Edu & Education & Primary School Teacher, University Professor, Vocational Trainer, Special Education Teacher \\
Heal & Human Health and Social Work Activities & Doctor, Nurse, Physical Therapist, Psychologist \\
Arts & Arts, Entertainment, and Recreation & Actor, Musician, Artist, Writer \\
\bottomrule
\end{tabularx}
}%
\end{table}

Personality traits are modeled using the Myers-Briggs Type Indicator (MBTI)~\cite{article}, a widely adopted psychological framework that categorizes individuals into 16 types across four dichotomous dimensions. In this work, we consider the eight individual MBTI traits: Extraversion (E), Introversion (I), Sensing (S), Intuition (N), Thinking (T), Feeling (F), Judging (J), and Perceiving (P). These attributes collectively define the persona, influencing the user simulator’s goals, dialogue behavior, and interaction patterns.

These attributes guide the user simulator’s goals, dialogue strategies, and behavioral patterns during interaction with the sales agent.

\subsection{Persona Dataset Formation}
While the personality dimension based on MBTI was derived from previous work, this study focuses on the remaining three attributes: gender, age, and occupation, treating personality as an independent variable.

To construct the persona dataset, we adopt a partial random sampling strategy: one attribute (e.g., gender) is fixed, while the remaining three attributes (age, occupation, and personality) are randomly sampled. For example, if gender is fixed, the corresponding age group, occupation, and MBTI personality type are randomly assigned to each simulated user.

For each fixed attribute condition, we generate 20 unique user personas. Persona generation is guided through structured prompting of LLaMA-3.1-8B-Instruct~\cite{llama3}, enabling coherent and diverse persona instantiation aligned with the specified constraints.


\subsection{Simulation}

Each simulation involved a dialogue between two models: a user simulator and the \textsc{SalesAgent}. The user simulator was responsible for generating persona-driven user utterances, while the \textsc{SalesAgent} produced both internal reasoning (\textit{thought}) and outward responses at each turn.

For every generated user persona, 15 conversations were conducted. Each conversation proceeded in a turn-based fashion, with the following termination conditions:

\begin{itemize}
    \item The conversation reached the predefined maximum of 20 turns.
    \item The sales agent explicitly ended the interaction with the utterance ``bye''.
    \item The thought of the sales agent concluded with the statement: ``The user has explicitly shown his/her intent of \texttt{<intent>}.''
\end{itemize}

\begin{figure}[t!]
\centering
\includegraphics[width=0.85\linewidth]{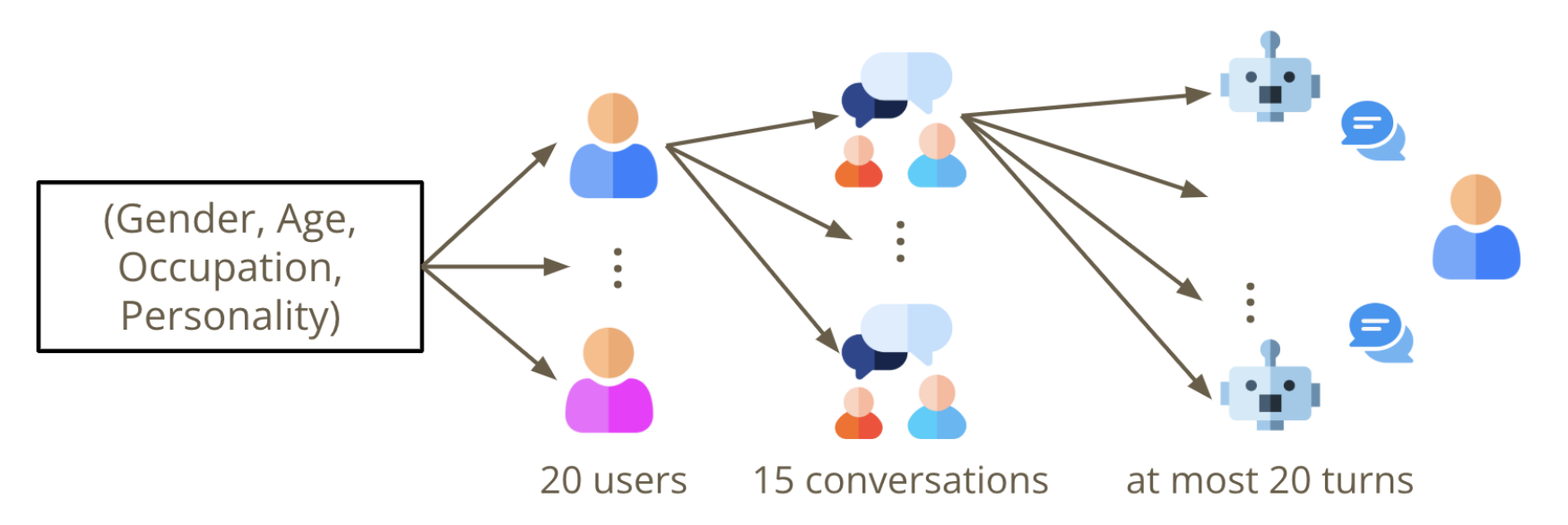} 
\vspace{-2mm}
\caption{Simulation overview.}
\label{fig:simulation}
\end{figure}

A conversation was considered successful if the sales agent's internal thought at termination was ``The user has explicitly shown his/her intent of \texttt{<intent>}.'' The simulation overview is illustrated in Fig.~\ref{fig:simulation}.

\subsection{Metrics}

Inspired by Cheng et al.~\cite{cheng2025exploringpersonalityawareinteractionssalesperson}, two metrics were adopted to evaluate the performance of the sales agent, allowing assessment not only of the success rate but also the understanding of the interaction efficiency. We also constructed two visualizations to further analyze the user and the sales agent’s behavior across intents.

The \textbf{success rate} is defined as the proportion of conversations in which the sales agent successfully identifies the user’s intent out of the total number of conversations conducted.

The \textbf{average number of turns} specifically measures the mean number of turns for only those conversations that ended in success. This helps assess how efficiently the sales agent identifies user intent when it is ultimately successful.



Additionally, we construct an \textbf{intent distribution} to capture how conversation topics evolve across interactions. Since multiple intents may appear within a single conversation, each intent is inferred from the agent’s thought annotations, with consecutive occurrences of the same intent counted as a single instance. The frequency of all identified intents is then aggregated across conversations to generate the overall distribution.
Building on this, we also derive a \textbf{success intent distribution}, which focuses specifically on intents that lead to successful outcomes. For each successful conversation, we extract the agent’s final thought to identify the concluding intent, and aggregate these across all successful cases to obtain the success intent distribution.


\section{Persona Attribute Analysis}

Simulations were conducted for each user attribute—age, gender, and occupation—resulting in a total of 9000 conversations and at most 180,000 dialogue turns. The user model was LLaMA-3.1-8B-Instruct~\cite{llama3}, and the agent model was \textsc{SalesAgent}. The resulting metrics are summarized in Table~\ref{tab:exp1_metrics}. In the following subsections, we discuss our findings with respect to age, gender, and occupation.

\begin{table}[t]
\centering
\footnotesize
\caption{SalesAgent performance when interacting with different user types. $^\ddagger$: $p < 0.05$; $^\dagger$: $p < 0.10$ in significance tests.}
\label{tab:exp1_metrics}
\begin{tabular}{llcc}
\toprule
& \textbf{Attribute} & \textbf{Success Rate} & \textbf{Avg. \#Turns ($\downarrow$)} \\ 
\midrule
\multirow{4}{*}{\it Age} & Teen    & 0.46 & \textbf{11.38}\\
& Adult   & \textbf{0.61}$^\ddagger$ & 11.61\\ 
& Middle-aged  & 0.57 & 11.96\\
& Elderly & 0.54 & 12.08\\ 
\midrule
\multirow{2}{*}{\it Gender} & Female & 0.50 & \textbf{11.34}\\
& Male   & \textbf{0.57} & 11.80\\ 
\midrule
\multirow{6}{*}{\it Occupation}& \textbf{Agr}iculture & \textbf{0.57}$^\dagger$ & 11.69\\ 
& \textbf{Info}rmation & 0.55 & 12.26 \\
& \textbf{Fin}ancial    & 0.51 & \textbf{11.63} \\
& \textbf{Edu}cation   & 0.53 & 11.84 \\
& \textbf{Heal}th & 0.50 & 12.00 \\
& \textbf{Arts}        & 0.51 & 12.35 \\ 
\bottomrule
\end{tabular}
\vspace{-2mm}
\end{table}

\subsection{Age}
Several insights emerge from the analysis of simulation metrics across different age groups. Among the four categories, adult users exhibited the highest success rate, while teens had the lowest. This trend may be attributed to differences in financial independence—adults, having disposable income, may be more susceptible to persuasive interactions than teens, who typically lack purchasing power. 

To assess the statistical significance of these patterns, we conducted an analysis of variance (ANOVA) across the four age groups. As shown in Table \ref{tab:exp1_metrics}, the success rate yielded a $p = 0.02$, indicating a statistically significant difference ($p<0.05$) in persuasion effectiveness among age groups.

In terms of the distribution of successful intents across age categories, no substantial differences were observed. In all four groups, \textit{FindRestaurant} consistently emerged as the dominant successful intent, suggesting that certain user goals remain universally popular regardless of age.
\subsection{Gender}
Across all metrics used to evaluate the outcomes of our simulation, we observed no significant differences between male and female user personas. The only notable deviation was in the success rate, where male users achieved higher scores than their female counterparts—contrary to our initial hypothesis. A possible explanation for this discrepancy is that the \textsc{SalesAgent} primarily focuses on detecting user intent rather than facilitating the actual act of purchasing. To assess whether the observed difference was statistically significant, we conducted independent two-sample t-tests for each metric. For the success rate, despite the apparent difference, the resulting $p$ was $0.15$. This suggests that the difference is not statistically significant at the conventional 0.05 threshold, indicating that gender may not be a dominant factor influencing user intent or behavior in this simulation framework. 
Furthermore, the distribution of successful intents across genders reveals a high degree of similarity. 

For both male and female users, \textit{FindRestaurants} and \textit{FindEvents} consistently emerged as the two most frequently successful intent categories. This further supports the observation that gender does not substantially influence the agent’s persuasive effectiveness or the users’ intent preferences in our setting.
\subsection{Occupation}

The analysis of role-playing simulations across occupational groups reveals notable differences in success rate. Users from the Agriculture, Forestry, and Fishing sector (\textit{Agr}) exhibited the highest success rate (0.57), suggesting that \textsc{SalesAgent} was more effective at guiding these users toward explicit intent expression.

\begin{figure}[t!]
\centering
\includegraphics[width=0.85\linewidth]{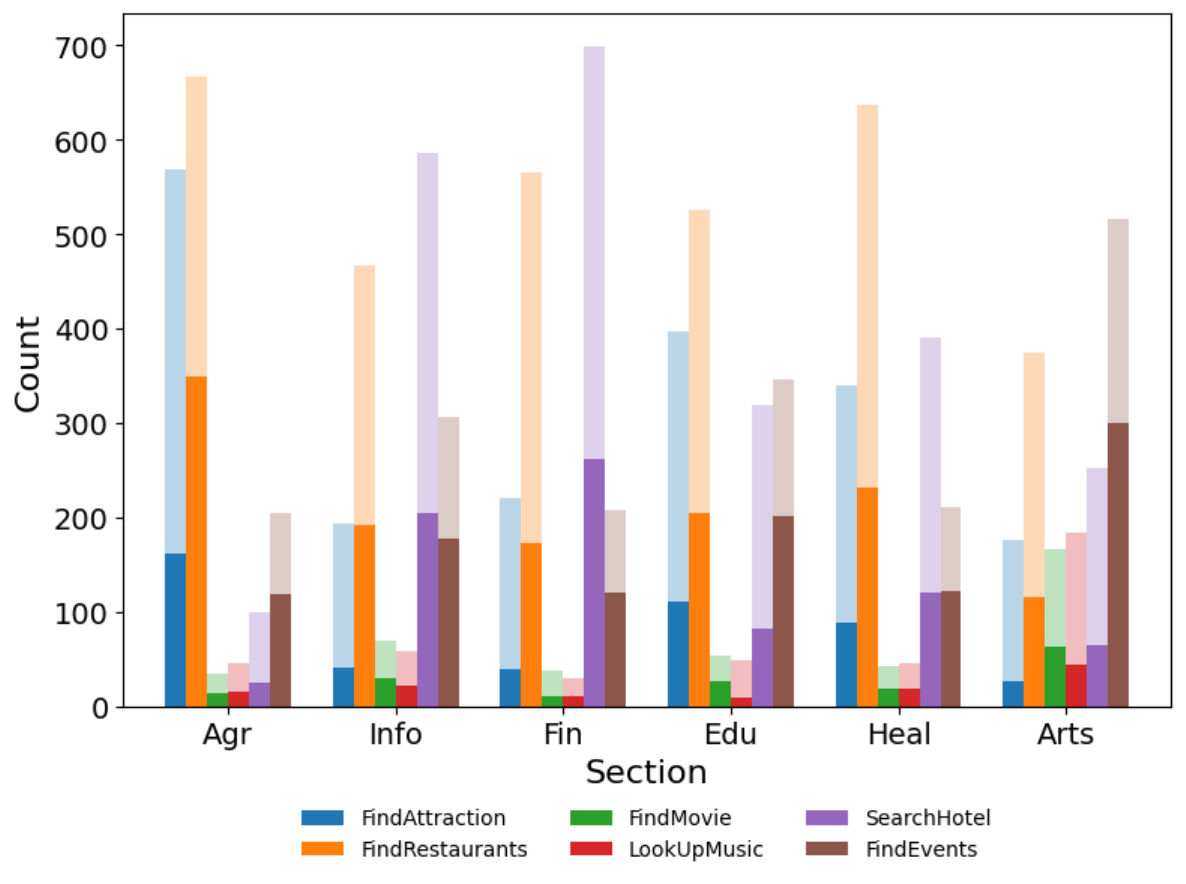}
\caption{Comparison of intent and success intent distributions across different user occupations in \textsc{SalesAgent}. Transparent bars represent overall intent distributions, while solid bars indicate success intent distributions.}
\label{fig:exp1_occupation_intent_comparison}
\end{figure}

According to Fig.~\ref{fig:exp1_occupation_intent_comparison}, \textsc{SalesAgent} is capable of inferring the user's occupation and adapting to varying intent distributions during simulation. The figure also reveals a positive correlation between intent frequency and success rate—intents that occur more frequently are also more likely to lead to successful outcomes. This suggests that users with different occupations tend to exhibit distinct intent preferences. For instance, users with recreational occupations (\textit{\textbf{Arts}}) tend to be more interested in \textit{FindEvents}; users with finance-related occupations (\textit{\textbf{Fin}}), in contrast, are more likely to reveal intents of \textit{SearchHotel}. To validate this observation, we conducted an analysis of variance (ANOVA) on the relationship between occupation sections and intent distributions. The resulting $p < 0.01$ indicates a statistically significant difference.

\begin{table}[htbp]
\centering
\footnotesize
\caption{Mapping of strategy prompts for different user occupations.}
\label{tab:occupation_strategy}

\resizebox{0.85\linewidth}{!}{%
\begin{tabularx}{\linewidth}{c p{1.9cm} X}
\toprule
\textbf{Sec.} & \textbf{Top-2 Intents} & \textbf{Rationale} \\
\midrule
Agr & FindRestaurants, FindAttraction & These users often value relaxation and leisure experiences when off work. \\
\midrule
Info & SearchHotel, FindRestaurants & Tech workers frequently travel for work and value reliable accommodations and good dining options. \\
\midrule
Fin & SearchHotel, FindRestaurants & These users may have business travel needs and typically prefer higher-end services. \\
\midrule
Edu & FindRestaurants, FindEvents & Educators often enjoy social or cultural activities and group-friendly dining. \\
\midrule
Heal & FindRestaurants, FindEvents & These users often seek stress relief through leisure activities and social events. \\
\midrule
Arts & FindEvents, FindRestaurants & Creatives are usually interested in events and venues that provide inspiration or entertainment, along with unique dining experiences. \\
\bottomrule
\end{tabularx}%
}
\end{table}

\begin{figure}[t!]
\centerline{\includegraphics[width=.7\linewidth]{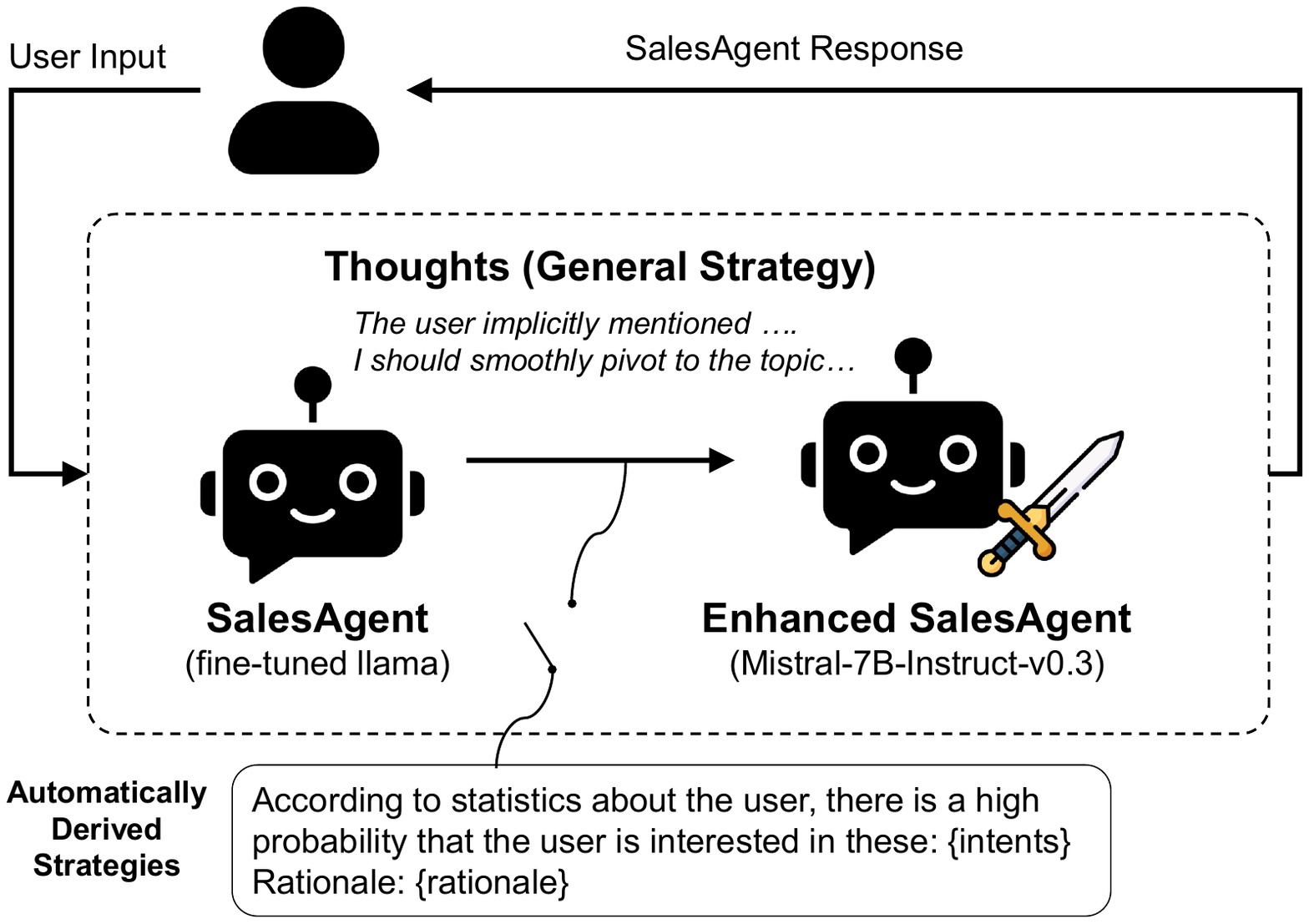}}
\vspace{-3mm}
\caption{Pipeline of \textsc{SalesAgent} w/ or w/o the occupation-based strategies.}
\label{fig:strategy-pipeline}
\end{figure}

\section{Occupation-Based Strategy for \textsc{SalesAgent}}

Building on insights from the previous experiment—which revealed pronounced differences in intent distributions across user occupations—we aimed to test whether such differences could be leveraged to create an effective, personalized dialogue strategy. Specifically, this experiment investigates whether a simple strategy, automatically derived from simulated user behaviors, can generalize to improve agent performance when applied to interactions with unseen user simulators.

To this end, we designed a lightweight occupation-based strategy in which \textsc{SalesAgent} prioritizes the two intents most frequently associated with each occupation, as identified through prior simulation analysis (Table~\ref{tab:occupation_strategy}). The goal was to evaluate whether this automatically derived strategy meaningfully improves success rates and conversational efficiency when applied to new user interactions, under the assumption that the user’s occupation is known.

To validate the generalization of this approach, we deliberately used different models for simulation and testing. The strategy was derived from intent distributions observed in the first experiment (using one simulator), but in this experiment, user inputs were generated by Qwen3-8B~\cite{qwen3technicalreport}, a different LLM, to ensure that the strategy was not overfitted to a particular simulator for generalizability. This design tests whether patterns learned from one simulated population can transfer to interactions with different user models.

In addition to testing strategy effectiveness, we sought to validate the modularity of this approach. We implemented a lightweight pipeline where \textsc{SalesAgent} provides high-level dialogue planning (i.e., thought generation), while a vanilla LLM generates the final user-facing response. Specifically, Qwen3-8B simulates user input, \textsc{SalesAgent} generates intermediate thoughts, and Mistral-7B-Instruct-v0.3~\cite{jiang2023mistral7b} produces the final response conditioned on both the thought and the strategy input. This architecture highlights how strategy augmentation can be flexibly applied to off-the-shelf LLMs.

This experiment focused specifically on occupation as the differentiating variable, with other user attributes (age, gender, MBTI) held constant across conditions. All other experimental settings, including the number of conversations and termination criteria, remained consistent with the prior study. The overall pipeline is illustrated in Fig.~\ref{fig:strategy-pipeline}.

\begin{table}[t!]
\caption{Performance of the occupation-based SalesAgent (w/o vs. w/ strategies).}
\label{tab:occupation_based_salesagent_metrics}
\centering

\resizebox{0.85\linewidth}{!}{%
\begin{tabular}{c c c c}    
\toprule
\textbf{Sec.} & \textbf{Success Rate} & \textbf{Avg. \# Turns ($\downarrow$)} & \textbf{Guided Conti. Ratio} \\
\midrule
Agr & 0.19 / \textbf{0.40} & 18.08 / \textbf{15.60} & \textbf{0.67} / 0.63 \\
Info & 0.27 / \textbf{0.35} & 17.05 / \textbf{16.38} & 0.65 / \textbf{0.66} \\
Fin & 0.23 / \textbf{0.36} & 17.41 / \textbf{16.27} & \textbf{0.71} / 0.62 \\
Edu & 0.21 / \textbf{0.74} & 17.70 / \textbf{10.96} & \textbf{0.71} / 0.51 \\
Heal & 0.15 / \textbf{0.61} & 18.26 / \textbf{13.23} & \textbf{0.74} / 0.55 \\
Arts & 0.26 / \textbf{0.68} & 17.40 / \textbf{11.85}  & \textbf{0.61} / 0.55 \\
\bottomrule
\end{tabular}%
}

\end{table}

\begin{figure*}[htbp]
\centering

\resizebox{0.85\linewidth}{!}{%
\begin{minipage}{\linewidth}
\centering
\subfigure[Without strategy]{
    \includegraphics[width=0.48\linewidth]{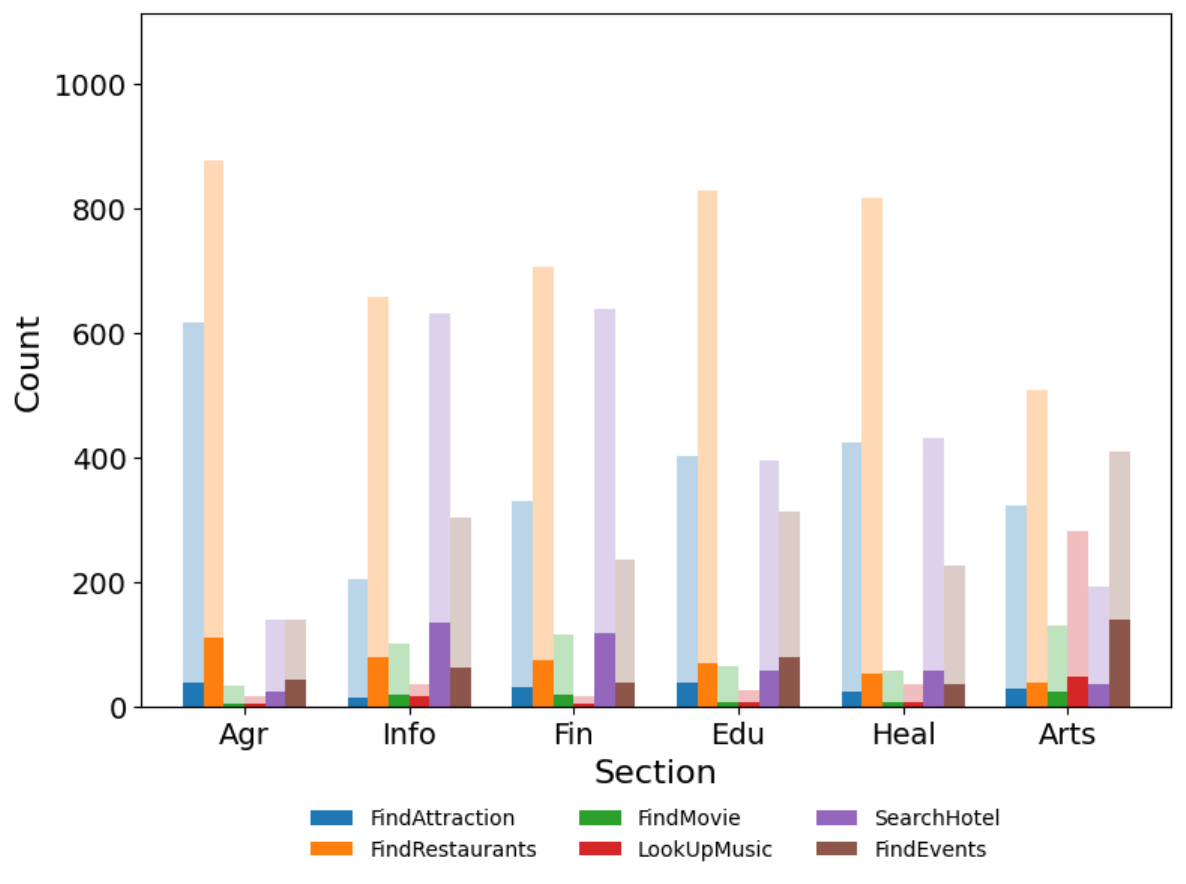}
    \label{fig:no_strategy_intent_dist}
}
\subfigure[With strategy]{
    \includegraphics[width=0.48\linewidth]{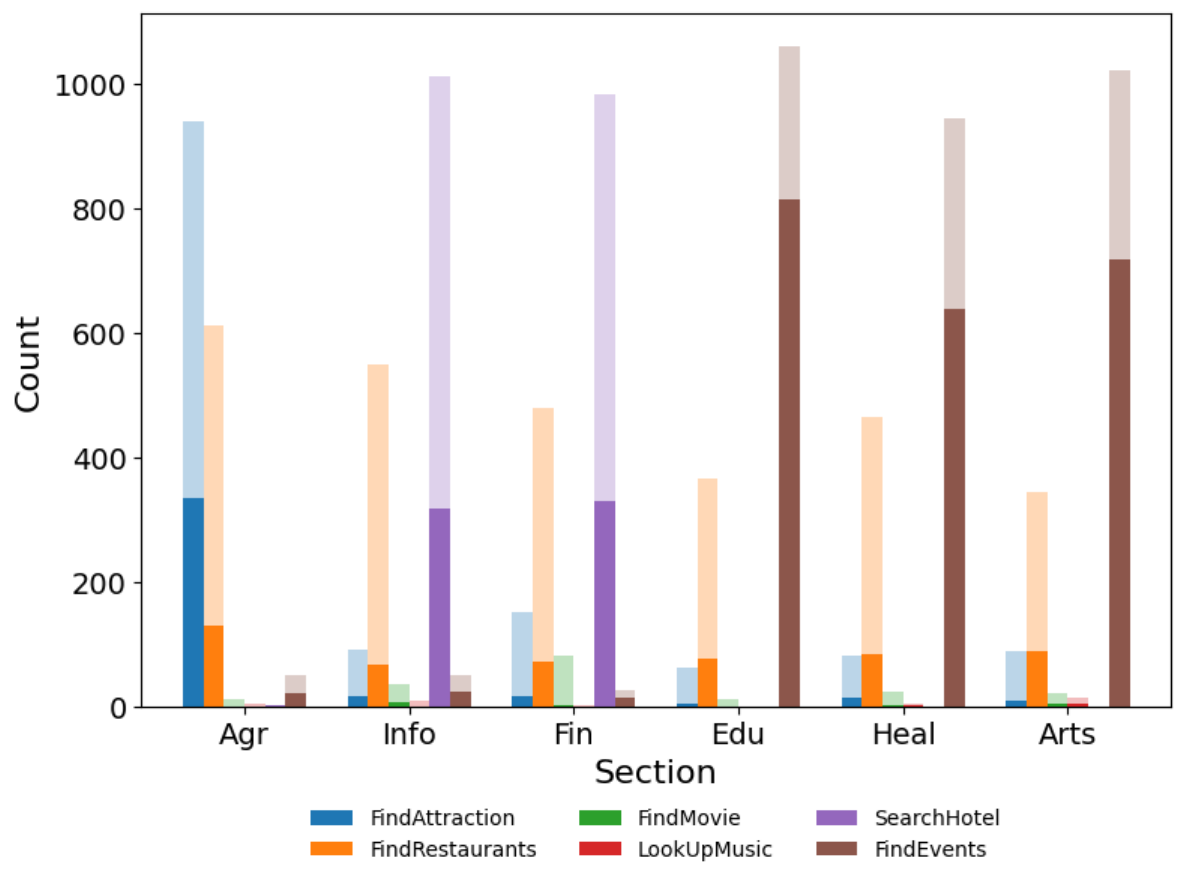}
    \label{fig:strategy_intent_dist}
}
\end{minipage}
}%

\caption{\textsc{SalesAgent}'s intent distributions and success intent distributions w/o vs. w/ strategies when interacting with unseen users in the testing phase. Transparent bars represent overall intent distributions, while solid bars indicate success intent distributions.}
\label{fig:strategy_vs_no_strategy_intent}
\vspace{-2mm}
\end{figure*}

\medskip
\noindent \textbf{Additional Metric.}
In this experiment, we introduced another metric, \textbf{guided continuation ratio}, a slight modification of the continuation ratio proposed by Cheng et al.~\cite{cheng2025exploringpersonalityawareinteractionssalesperson}. This metric measures the proportion of the sales agent’s internal thoughts labeled as “continue” that occur immediately after the point at which the agent decides to pivot the conversation toward a specific topic. This metric helps evaluate the aggressiveness of the agent: the higher the ratio, the more aggressive the agent is.

\medskip
\noindent \textbf{Results and Insights.}
Table~\ref{tab:occupation_based_salesagent_metrics} demonstrates that the occupation-based strategy yielded notable improvements in success rate, with sections \textit{Edu}, \textit{Heal}, and \textit{Arts} showing gains exceeding 40\%. Moreover, both the guided-continuation ratio and the average number of turns declined across most occupations, for both successful and unsuccessful dialogues. This suggests that the strategy enables the agent to more effectively assess user interest and conduct conversations with greater precision. At the same time, however, the results also indicate that the strategy markedly increases the agent’s aggressiveness, highlighting a \textbf{trade-off between higher success rates and conversational aggressiveness.}

Fig.~\ref{fig:strategy_intent_dist} further illustrates how the strategy steers conversations toward intents that align with the user’s occupation and correlate with higher success rates. Compared to the no-strategy condition (Fig.\ref{fig:no_strategy_intent_dist}), the agent guided by the strategy significantly reduces conversation drift toward irrelevant intents and achieves a higher overall success rate. These findings reinforce the effectiveness of personalized targeting based on user occupation.
Importantly, when comparing Fig.\ref{fig:exp1_occupation_intent_comparison} (from our initial experiment) with the current no-strategy baseline (Fig.\ref{fig:no_strategy_intent_dist}), we observe consistent intent distributions across occupational groups, despite differences in user simulation models. This consistency supports the robustness of the observed occupation-based intent preferences and demonstrates that the strategy derived from one simulator can generalize to new user populations.

\medskip

\noindent \textbf{Implications.}
These results demonstrate that a simple, simulation-derived occupation-based strategy can significantly enhance sales-oriented dialogue outcomes, even when applied to interactions with unseen user simulators and combined with generic LLMs. The approach offers a lightweight and flexible framework for future deployment: personalized strategies can be dynamically composed based on user attributes and seamlessly integrated with off-the-shelf LLMs, with \textsc{SalesAgent} serving as a modular strategy planning layer. This paves the way for scalable, occupation-sensitive dialogue systems capable of adapting to diverse user populations without extensive retraining.




\section{Limitations and Future Work}

While our proposed strategy significantly improves the success rate of the sales agent, several limitations remain. First, both the user and the sales agent are simulated by language models rather than real users, which may not fully capture the complexity, variability, and unpredictability of real human behavior. Second, the internal thoughts generated by \textsc{SalesAgent} are not always accurate—for instance, the model may misinterpret the user’s intent, resulting in responses that are misaligned with the conversational context.

Moreover, the current strategy is relatively coarse, as it only utilizes occupation-level preferences. A more fine-grained approach, such as incorporating individual personality traits (e.g., MBTI types), could offer deeper personalization. The agent also shows limited adaptability: when a user appears disinterested in a given topic, the agent does not proactively explore alternative topics, potentially reducing persuasion effectiveness.

Another constraint is that our occupation-based strategy currently depends on explicitly provided occupation types and their corresponding strategies. Future work could involve developing or fine-tuning an in-context learning mechanism for dynamically detecting a user’s occupation or personality type (e.g., MBTI) through interaction turns, and applying tailored strategies accordingly.

Also, although the occupation-based strategy achieved a higher success rate, the results also indicate an increase in the agent’s aggressiveness. Future work could investigate reinforcement learning–based approaches to better balance success rate with conversational aggressiveness.

Finally, certain user personas—especially those with passive or “quiet” characteristics—pose a significant challenge, as they are generally less responsive and harder to persuade regardless of strategy. Improving the agent’s ability to detect and adapt to such personas remains an important direction for future research.

\section{Conclusion}

In this study, we conducted a comprehensive analysis of how user personas—specifically gender, age, and occupation—influence the conversational dynamics of a sales-oriented agentic dialogue system. Our investigation revealed distinct patterns in how these persona attributes affect key outcomes, including success rate, number of turns, and intent distributions.

Among these attributes, occupation emerged as the most informative factor for strategy design. Leveraging insights derived from user-simulator experiments, we constructed lightweight, simulation-informed strategies that encode occupation-level preferences into an LLM-based sales agent. Importantly, this approach requires no model fine-tuning, yet it achieves significant improvements in success rate through prompt-level adjustments and strategy conditioning. By testing the strategy on interactions with unseen simulators from a different model, we demonstrate its generalization capability. These findings highlight both the effectiveness and transferability of persona-informed approaches for enhancing persuasive dialogue systems. We believe that such light-weight framework can serve as a starting point of dialogue customization, paving the way for a more adaptive and user-centric system.
\section{Acknowledgements}

This work was financially supported by the National Science and Technology Council (NSTC) in Taiwan, under Grants 111-2222-E-002-013-MY3 and 112-2223-E002-012-MY5. 
We thank the National Center for High-performance Computing of National Institutes of Applied Research (NIAR) in Taiwan and Google's PaliGemma Academic Program
for providing computational and storage resources.


\bibliographystyle{IEEEtran}
\bibliography{references} 

\appendices

\end{document}